# A knowledge-driven framework for synthesizing designs from modular components

Constantin Chaumet[a,*], Jakob Rehof[a], Thomas Schuster[a]

[a]Software Engineering by Algorithms and Logic, Department of Computer Science, TU Dortmund University, Otto-Hahn-Str. 12, 44227 Dortmund, Germany

* Corresponding author. Tel.: +49-231-755-7719; fax: +49-231-755-7936. E-mail address: constantin.chaumet@tu-dortmund.de

**Abstract**

Creating a design from modular components necessitates three steps: Acquiring knowledge about available components, conceiving an abstract design concept, and implementing that concept in a concrete design. The third step entails many repetitive and menial tasks, such as inserting parts and creating joints between them. Especially when comparing and implementing design alternatives, this issue is compounded. We propose a use-case agnostic knowledge-driven framework to automate the implementation step. In particular, the framework catalogues the acquired knowledge and the design concept, as well as utilizes Combinatory Logic Synthesis to synthesize concrete design alternatives. This minimizes the effort required to create designs, allowing the design space to be thoroughly explored. We implemented the framework as a plugin for the CAD software Autodesk Fusion 360. We conducted a case study in which robotic arms were synthesized from a set of 28 modular components. Based on the case study, the applicability of the framework is analyzed and discussed.

## 1. Introduction

The emergence of Industry 4.0 has shifted products towards being highly customizable goods [1] and being able to offer customized designs is a significant competitive advantage [2], as long as the price of the final product does not increase [1]. For surface level customizations of designs this is often the case, e.g., freely customizing the color of a design. However, deeper customizations (physical structure, internal hardware), often necessitate many further design changes. Implementing such dependent changes requires time investment, increasing the price.

This issue is not limited to customization. During product development, a design is often iteratively refined from rough ideas [3]. Lessons learned from prior design stages are applied to improve the overall design. A significant portion of the total development time is implementing, examining, and testing design alternatives in CAD software. When the design is well modularized, the implementation step is straightforward, albeit menial and repetitive, as certain modular components are shared between most design alternatives.

There have been numerous previous approaches to increase automation of that step, such as high-level CAD templates [4], probabilistic synthesis from components [5], linkage graphs obtained by machine learning [6], semi-automated recommender systems based on probabilistic factor graphs [3], web-based KBS systems [7], and automated varying and filtering of existing parametric designs [8]. There are also some methods that do not operate on CAD data, such as mesh-based machine learning [9,10] and volumetric machine learning [11], resulting in meshes as output format.

In this paper we propose a use-case agnostic knowledge-driven framework based on Combinatory Logic Synthesis [12] which fully automates the implementation step. Our approach differs from the existing work in three key aspects. Firstly, there is no probabilistic or machine learning component needed for the synthesis itself, results are guaranteed to conform to the knowledgebase. Secondly, our approach is fully compatible with pre-existing CAD data, regardless of design history, or if direct or parametric modeling was used. Finally, the input and output are CAD data files in widely used CAD software.

Since the effort of creating a design is minimized, and the speed of design creation is approximately two orders of magnitude faster than manual design, the framework encourages thorough design space exploration, as the opportunity costs are low. The framework is implemented in Autodesk Fusion 360 [13]. Other CAD software from Autodesk has been shown to be suitable for applying knowledge-based techniques [14].

## 2. Combinatory Logic Synthesis

Combinatory Logic Synthesis is a technique that composes modular components contained in so-called repositories based on types [15]. The types define valid compositions of the components, as well as encoding domain specific knowledge. Automated synthesis of such valid compositions is performed by solving the type inhabitation problem.

The type inhabitation problem poses the following question: given a target type $\tau$ and a repository $\Gamma$, can a term (a well-typed composition of the typed components contained in $\Gamma$) be found that is of the type $\tau$? Formally, this is denoted as $\Gamma \vdash ? : \tau$.

$$M, N ::= c \mid (MN) \tag{1}$$
$$\sigma, \tau ::= a \mid \sigma \to \tau \mid \sigma \cap \tau \tag{2}$$

Equation 1 states that any term is either a singular component from the repository, $c$, or an application of the term $M$ to $N$. Equation 2 describes how components contained in the repository can be typed. The simplest type of a component is $a$, which assigns a semantic identifier. These simple identifiers can be composed into more complex types. Functional types of the form $\sigma \to \tau$ can take an input value of type $\sigma$ and return values of type $\tau$. Intersection types of the form $\sigma \cap \tau$ describe a *logical and* between $\sigma$ and $\tau$, any component of this type must be a valid result to a request of $\sigma$ as well as $\tau$. By combining functional and intersection types, components can be precisely specified w.r.t. to the functionality they provide and which dependencies they require.

The automated synthesis also takes into consideration a taxonomy (usually a hierarchical ordering) of the types present in the repository. Taxonomies allow generalizing types, e.g., a *"servomotor"* can be a subtype of *"motor"*. Any request for a motor can also be satisfied by one of its subtypes, i.e., *"servomotor"*. Details on how this interacts with functional and intersection types can be found in [16].

The technique has been shown to be applicable to several real world problems, including automatic generation of factory simulation models [17], motion planning programs for robotic systems [18], as well as automated CAM toolpath generation [19]. There are libraries available to leverage Combinatory Logic Synthesis in the Scala and Python programming languages [20,21].

## 3. Framework

In this section the framework itself is covered in detail. First, an overview of the architecture is presented. Then, the intended workflow to prepare and synthesize designs is explained on a high level. Finally, the individual steps are covered in more detail, explaining methodology and corresponding implementation details.

### 3.1. Architecture

The framework is split into a web-based backend and a Fusion 360 plugin as a front-end. Each of these is split into several modules. The code is released publicly under the Apache License 2.0 at [22].

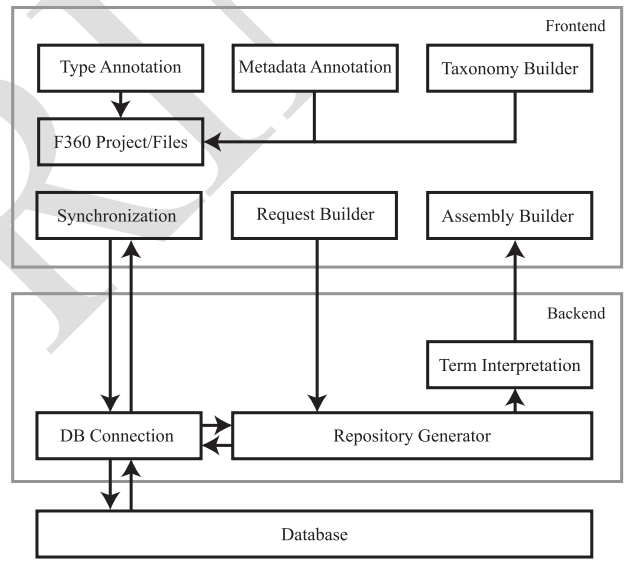

Fig. 1. Schematic overview of framework architecture.

Figure 1 illustrates the modules present in the front- and backend and their interactions, respectively. With exception of the synchronization and F360 Project/Files modules, all modules in the frontend are implemented by means of responsive graphical user interfaces natively integrated into the CAD software. The taxonomy builder module visually displays the current taxonomies and allows editing them freely. The type annotation module allows traversing, selecting, and intersecting types from taxonomies and annotating these to CAD geometry or entire files. The metadata annotation module allows adding additional information to files, i.e., costs or projected availability of a mechanical component. These three modules serve to encode the domain-specific knowledge, further details on this are given in Section 3.3. All the knowledge encoded this way is persisted in the project and files, as well as continually synchronized with a MongoDB database. The request builder module allows specifying an inhabitation request problem. Users graphically build a request by specifying the kind of design to create and selecting additional conditions. The module then translates this into a formal request that the Combinatory Logic Synthesis can process.

The main module of the backend is the repository generator. It receives the requests created in the frontend and fetches the necessary knowledge (types, modular components, metadata) from the database. From this retrieved data, the repository for the Combinatory Logic Synthesis is dynamically constructed. A detailed explanation is given in Section 3.4. After the repository generator has constructed a repository based on the request, the request is processed. The result is a set of terms that represent design solutions. The terms are highly nested sequences of applications as described in Equation 1. The term interpretation module processes these into a flattened list of assembly instructions, forming an assembly program that the assembly builder module in the frontend can preview or execute. Previewing a program gives a BOM of the design and displays aggregated metadata, executing it creates the design as a fully functional assembly in the CAD software.

*3.2. Workflow*

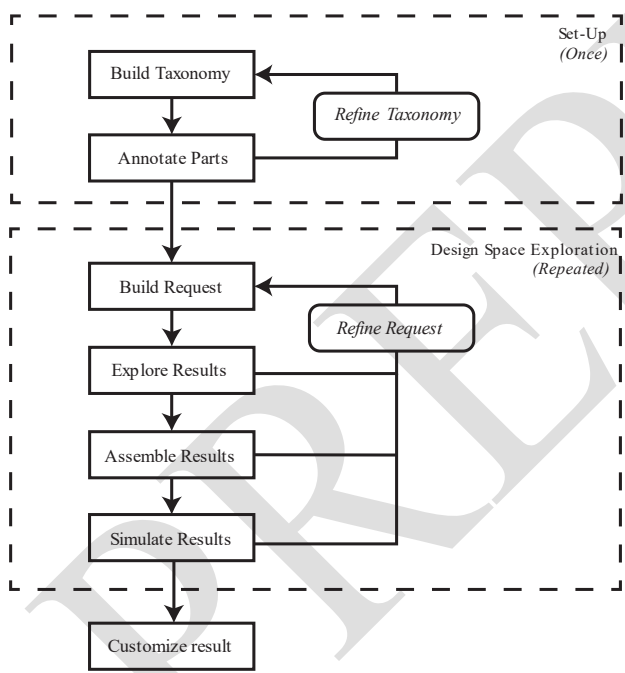

Fig. 2. Workflow to synthesize designs.

Figure 2 illustrates the basic workflow of synthesizing designs. The workflow is split into two distinct phases, the initial set-up, and the following repeated design space exploration. The set-up phase can be time-intensive, depending primarily on the number of modular components that need to be prepared. However, this time-intensive set-up only needs to be carried out once and is usually quickly amortized during the design space exploration phase. The design space exploration phase has a simple iterative structure. First, an initial request is built, describing the design to synthesize. Initial requests may be relatively unspecific (i.e., asking for a goal type high up in the taxonomy) and may therefore yield a relatively large set of designs as solutions. The resulting designs are then examined in an abstract representation and the user may browse a list which contains key metrics of the designs, i.e., the BOM. The performance is very good, the set of results only requires a few seconds to be generated. Based on this initial examination, the user either refines the query to be more specific or proceeds to the next step.

In this next step, the user can elect to either assemble individual or a batch of results in the actual CAD software. Assembly is more time-consuming than the previous step (per examined result) but is still orders of magnitude faster than a human, as shown by the case study in Section 4.

*3.3. Set-Up Phase*

The set-up phase is an iterative process consisting of two steps. The user starts by building an initial taxonomy using the taxonomy builder module. The initial taxonomy does not need to be exhaustive, but should already contain most abstract categories, the "top level". The taxonomy can be split into an arbitrary set of smaller and more specific taxonomies to reduce the individual complexity. The framework internally merges these taxonomies into a combined one. After the user has created an initial taxonomy, containing general types such as *"screw", "metric", "steel",* the modular components can then be enriched with type information. For each component, the user marks intended connection points, and assigns them a coordinate system and a joint type (either revolute or rigid). Our plugin for the CAD software enables the user to construct these coordinate systems from reference geometry. The user then assigns either a required or a provided type to the coordinate system, or both, using the type annotation module. The assigned type is usually an intersection of several types from the taxonomies, for instance, a connection could be typed as requiring a metric steel screw by setting the required type of a coordinate system to the intersection of the types *"metric", "steel",* and *"screw"*.

In addition to the coordinate systems, the component itself is also assigned a type which describes its inherent attributes, for instance a ceramic ball bearing might be typed by the intersection of *"bearing", "steel",* and *"ceramic".* Any coordinate systems assigned a provided type get additionally intersected with these inherent attributes.

It is important to not "overspecify" these intersection types. They should be kept as general as possible and accurately reflect the full scope of connection possibilities. For example, a threaded hole on a flat surface should not require specifically a screw if there are other feasible fasteners in the set of modular components.

While typing individual components the user will discover that he is missing types in the taxonomy, or that a connection requires a more specific connecting component. When such a case occurs, the user refines the taxonomy by adding an appropriate new type or subtype. Our plugin facilitates this through interactive graphical interfaces provided natively in the CAD software. Domain specific knowledge about the modular components and the way they connect and interact gets iteratively encoded during this step.

*3.4. Design Space Exploration Phase*

In the design space exploration phase, the user submits an initial inhabitation request that only contains the most important metrics, e.g., structural or material constraints. Constraints can be any discrete metric. Based on this request, a repository for Combinatory Logic Synthesis is then dynamically created.

We model physical modular components by adding dynamically typed components to the repository, so called *"combinators"*. We assume that each physical modular component has been annotated with typed connection points and a type that describes its function as previously detailed in Section 3.3.

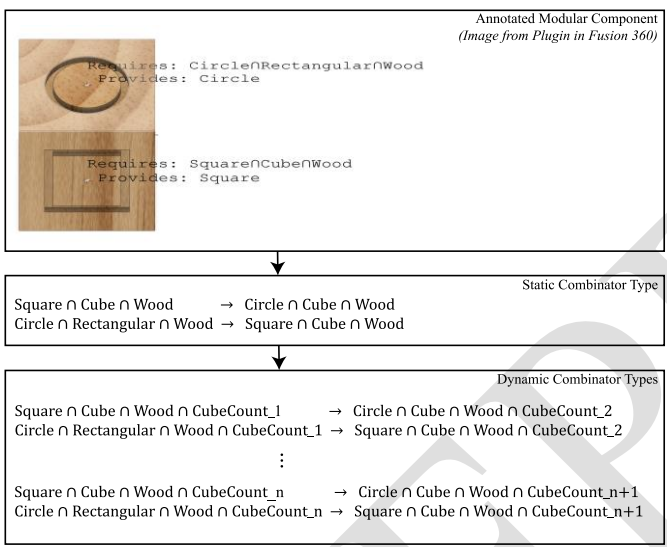

Fig. 3. Dynamic type generation for repository.

For each connection point that has been annotated with a provided type, an initial static type is generated for the component. Types are constructed from consecutive arrow types that take the required types of all other annotated connection points, and finally terminate in an arrow type that returns the provided type intersected with inherent types of the component. Figure 3 shows a simplified example with two connection points annotated with provided types. The shown component has the inherent properties *"Cube"* and *"Wood"*. As a result, two static types get generated as described.

Based on the inhabitation request, these static types get expanded to several dynamic types. These dynamic types allow aggregating information in the type system that is necessary to fulfill the request. In Figure 3, the user wants to request an assembly that contains a specific number of cubes in total. Dynamic types get generated that annotate all required types with the number of cubes that an attached subassembly could contain, and the returned type gets annotated with the sum of the number of cubes in all required subassemblies, incremented by 1, since the part itself is a cube. For components that have many different required parts, and if multiple properties are relevant to the inhabitation request, the number of dynamic types generated is the cartesian product of the possible values of the properties and the required types, resulting in many combinators being added to the repository. However, the inhabitation algorithm efficiently handles this, with the run-time required being usually negligible in comparison to the time to assemble results in CAD software.

After the inhabitation request is completed, the user can open an interface in our plugin to browse the results and evaluate them briefly based on the BOM. Results can also be selected for assembly in the CAD software. Thus, the user can sample the design space and note flaws or desirable properties of the resulting assemblies. Taking these into account, the user then narrows the design space by refining his inhabitation request with these desirable properties. This process is repeated until the results are sufficiently narrowed down.

Then, all remaining results are batch-assembled in the CAD software. Our plugin is implemented in such a way that there is a significant performance gain when multiple results get assembled at once. The user can then take an in-depth look at these results and run simulations or other testing measures. If this step uncovers some previously overlooked property, the inhabitation request gets refined, and the process is repeated. Else, the user picks one or several of the results and applies finishing touches or further customization.

## 4. Case Study

We conducted a case study to evaluate the applicability and performance of our framework. To this end, we designed a set of 28 modular components from which robotic arms can be constructed. 14 of these are off-the-shelf parts, consisting of servomotors and screws. The taxonomy and the typing of the parts was constructed as described in Section 3.1, taking a total of about 90 minutes to complete by a user familiar with combinatory logic. We added the approximate costs as metadata to the modular components based on online retailer prices for the off-the-shelf parts and used our estimated production costs for SLS printing for the remaining parts. The complete dataset including the taxonomy used is available at [23].

Parts that are not off-the-shelf components are designed to be manufactured through rapid prototyping technologies. We tested their manufacturability on an FDM printer using PLA, on an SLA Printer using UV resin, and on an SLS printer using Nylon 12. We encountered no issues with any of the parts during manufacturing, except for the gripper component, which is difficult to manufacture on FDM printers due to its complicated geometry requiring a lot of support structures during printing.

We synthesized and assembled all robotic arm designs of four (26), five (82) and six (256) degrees of freedom (DoF) in the CAD software using our plugin. We inspected the assembled robotic arms for interferences, adherence to the taxonomy and type annotations, or other features that might render them non-constructable. All robotic arms inspected showed no issues.

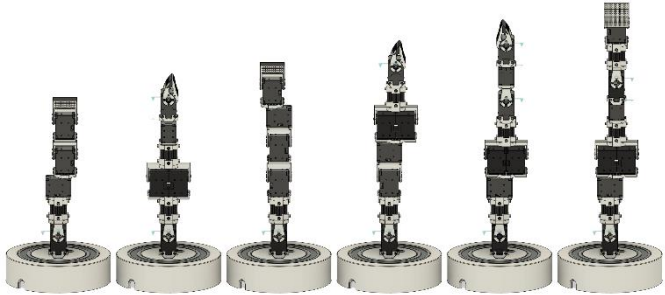

Fig. 4. Robotic arm designs assembled in CAD software by our framework.

Some of the assembled robotic arm designs are shown in Figure 4, two representatives of each of the inspected degree of freedom groups are pictured. During assembly of the designs, our plugin also manages the assembly tree, creating subassemblies for each separate kinematic link, and creating labelled groups for repeated parts like screws. Interacting with and customizing the synthesized assemblies is easier as a result.

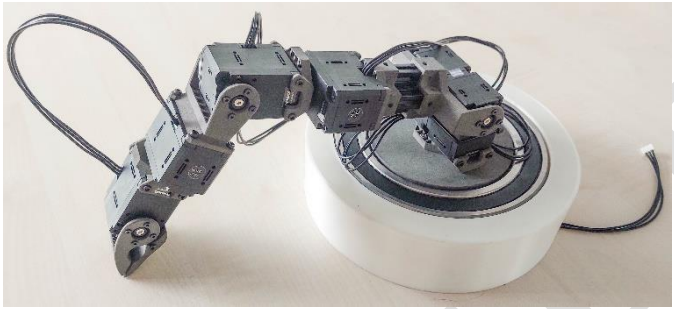

Fig. 5. Assembled robotic arm.

To verify that the robotic arms do not contain any difficult to spot issues caused by the synthesis or assembly processes, we randomly selected one of the results and assembled it from SLS-printed parts (Nylon 12). The assembled arm is pictured in Figure 5. There were no issues during the assembly process. Experimental verification that the assembled design is mechanically and kinematically sound is provided by a set of software packages that perform motion planning on the assembled robotic arm. We sampled and executed random valid poses for the robotic arm to assume. We observed correct behavior, the robotic arm assuming the poses in a collision free manner.

To quantify the performance of our plugin we compare it to that of a human user. First, we measured the time required for our framework to synthesize designs of different degrees of freedom. The beginning of the design space exploration phase is primarily impacted by this time, as the synthesis request is still being refined without assembling designs. Then, we measured the average time our framework takes to assemble single and multiple results. These values are distinct since our plugin can generate and use template files in the CAD software to be more efficient when assembling multiple results. The value for a single result is relevant to a designer fine-tuning the synthesis request and inspecting the assembled designs in depth. The value for multiple results is relevant for the final step of the design space exploration, where all remaining results are inspected, and final candidates are picked. The value for a single result was obtained by measuring the times needed to assemble the results with the smallest, largest, and median number of modular components, and averaging those times.

We then picked two representatives of each degree of freedom (Figure 4) and challenged a human user familiar with Fusion 360 to first construct one design of each degree of freedom from scratch. This value is intended to provide an estimate of the time that it takes a designer to create an initial draft of a design from a set of modular components (Table 1, "Human Single" column). We then asked the user to construct the other robotic arm design as efficiently as possible, reusing any parts of the previously assembled design. This value is intended to provide an estimate of the time that a designer requires to adapt a design, i.e., go through a design iteration (Table 1, "Human Multiple" column).

Table 1. Performance comparison between framework and human.

| DoF | Request Time | Framework Single | Framework Multiple | Human Single | Human Multiple |
|---|---|---|---|---|---|
| 4 | 0m 1s | 0m 52s | 0m 14s | 66m 5s | 42m 25s |
| 5 | 0m 2s | 0m 58s | 0m 16s | 59m 17s | 23m 11s |
| 6 | 0m 5s | 1m 03s | 0m 22s | 121m 28s | 32m 06s |

The results of this experiment can be seen in Table 1. Measurements were carried out on a workstation with an Intel Core i9-13900K processor. The request times for the robotic arms of the different degrees of freedom were negligible in comparison to all assembly times, human and plugin. The human results exhibit a lot of variance but still give a good estimate of the time required by a human for designs of this complexity. The times required by a human to modify one design to match another (last column) indicate that it is not primarily the number of components in a design that determines how fast an iteration can be created. Rather, the time is primarily determined by the complexity of the design and the number of labor-intensive changes needed to be made. The time spent by a human creating a fixed design from scratch gives an indication of how fast CAD software can be physically operated. Both metrics do not account for time spent thinking or on the creative process, which would cause more time to elapse. Both measured time categories represent typical use-cases when creating a design from modular components.

We exclude the time the user needed to familiarize himself with the set of components as well as the time it took to execute the set-up phase of our plugin, only comparing the iterative design exploration phases. Both times are up-front costs and are amortized as more designs get created from a given set of components.

We consider the values of the "Framework Single" and "Human Single" columns to be representative of the initial rough design space exploration stage, i.e., posing requests and assembling single results. Analogously, we consider the "Framework Multiple" and "Human Multiple" columns to be representative of the later design space exploration stage, where batches of remaining candidates get assembled and evaluated. In both stages our framework achieves about two orders of magnitude more efficient exploration of the design space than manual design. Additionally, our framework offers a more structured approach to exploring the present design space and makes no mistakes during assembly, for instance forgetting components that are occluded by other parts of the assembly.

## 5. Discussion and Outlook

Combinatory Logic Synthesis was used in conjunction with a plugin for the CAD software Fusion 360 to allow enriching sets of modular components with knowledge regarding their connectivity and purpose. This approach allows the knowledge to be reused and leverages it to synthesize designs. The proposed workflow facilitated by the plugin allows iteratively refining the encoded knowledge and the synthesized designs. The integration with CAD software allows finishing touches and customization to be part of the workflow, as the designs do not need to be exported and are in a native CAD file format, enabling large design spaces to be efficiently explored. Our case study shows that our approach allows for significant efficiency gains/time savings during an iterative design process. Unlike machine-learning based approaches, we can guarantee that generated designs conform to the knowledgebase.

In future work we intend to explore several augmentations to the proposed framework to achieve even greater increases in efficiency. One of these is augmenting the creation of taxonomies and assigning types by a machine learning based recommendation system. By doing this, the up-front time required to prepare a pre-existing set of modular components for use with our framework can be reduced. The exploration phase could also be accelerated by using black box optimization techniques. As evidenced in Section 3.4, the design space exploration phase is inherently iterative, and such lends itself to being part of an optimization loop. Similar to the approach used in [18], this can be leveraged to explore the design space in automated fashion, finding designs that are optimal with regard to the set of metrics the user is interested in.

Additionally, we intend to work on more comprehensive case studies, hoping to tackle use cases from industrial partners and benchmarking our framework's performance against a larger number of professional designers.

## Acknowledgements

This research was funded by the Deutsche Forschungsgemeinschaft (DFG, German Research Foundation – Project Number: 276879186).